\documentclass[letterpaper, 10 pt, journal, twoside]{ieeetran}

\usepackage{textcomp}
\usepackage{gensymb}
\usepackage{blindtext}
\usepackage{graphicx}
\usepackage{tikz}
\usepackage{tikz-3dplot}
\usepackage{mathtools}
\usepackage{amssymb}
\usepackage{mathbbol}
\usepackage{algorithm}
\usepackage[noend]{algpseudocode}
\usepackage[acronym]{glossaries} 
\usepackage[caption=false]{subfig}
\usepackage[all]{nowidow}
\usepackage{etoolbox}
\usepackage{threeparttable}
\usepackage{booktabs}
\usepackage{multirow}
\usepackage{adjustbox}
\usepackage{optidef}
\usepackage{cellspace}

\usetikzlibrary{shapes.geometric, arrows, positioning}
\tikzstyle{process} = [rectangle, rounded corners, fill=white, minimum width=2.5cm, text width=2.2cm, minimum height=1cm, text centered, draw=black]
\tikzstyle{arrow} = [thick,->,>=stealth]
\tikzstyle{stage} = [rectangle, rounded corners, fill=gray!40, minimum width=3cm, minimum height=5cm]

\definecolor{viridis0}{RGB}{68, 1, 84}    
\definecolor{viridis25}{RGB}{58, 82, 139} 
\definecolor{viridis50}{RGB}{32, 144, 240} 
\definecolor{viridis75}{RGB}{94, 201, 98}  
\definecolor{viridis100}{RGB}{253, 231, 36} 
\definecolor{planecol}{RGB}{32, 144, 240}
\definecolor{linecol}{RGB}{94, 201, 98}
\definecolor{ballcol}{RGB}{253, 231, 36}

\makeatletter
\AfterEndEnvironment{algorithm}{\let\@algcomment\relax}
\AtEndEnvironment{algorithm}{\kern2pt\hrule\relax\vskip3pt\@algcomment}
\let\@algcomment\relax
\newcommand\algcomment[1]{\def\@algcomment{\footnotesize#1}}

\renewcommand\fs@ruled{\def\@fs@cfont{\bfseries}\let\@fs@capt\floatc@ruled
	\def\@fs@pre{\hrule height.8pt depth0pt \kern2pt}%
	\def\@fs@post{}%
	\def\@fs@mid{\kern2pt\hrule\kern2pt}%
	\let\@fs@iftopcapt\iftrue}
\makeatother

\newcommand\eqrefb[1]{\mbox{(Eq. \ref{eq:#1})}}
\newcommand\algmref[1]{\mbox{(Alg. \ref{alg:#1})}}
\newcommand\secref[1]{\mbox{(Sec. \ref{sec:#1})}}
\newcommand\tsecref[1]{\mbox{Sec. \ref{sec:#1}}}
\newcommand\figref[1]{\mbox{(Fig. \ref{fig:#1})}}

\newcommand\eqlabel[1]{\label{eq:#1}}
\newcommand\algmlabel[1]{\label{alg:#1}}
\newcommand\seclabel[1]{\label{sec:#1}}
\newcommand\figlabel[1]{\label{fig:#1}}

\newcommand\tbllabel[1]{\label{tbl:#1}}

\usepackage[utf8]{inputenc}
\DeclareUnicodeCharacter{0229}{e}
\usepackage[
bibstyle=ieee,
citestyle=numeric-comp,
isbn=false,
doi=false,
eprint=false,
sorting=none,
url=true,
bibencoding=utf8,
autocite=plain,
maxcitenames=20,
maxbibnames=20,
backend=biber
]{biblatex}
\makeatletter
\renewbibmacro*{title}{%
	{\printtext[title]{\thefield{title}}}%
}
\makeatother

\DeclareFieldFormat{url}{\url{#1}}
\DeclareBibliographyCategory{needsurl}
\newcommand{\entryneedsurl}[1]{\addtocategory{needsurl}{#1}}
\renewbibmacro*{url+urldate}{%
	\ifcategory{needsurl}
	{\printfield{url}%
		\iffieldundef{urlyear}
		{}
		{\setunit*{\addspace}%
			\printurldate}}
	{}}

\title{EllipseLIO: Adaptive LiDAR Inertial Odometry with an Ellipsoid Representation}

\author{Rowan Border$\,^{1}$ 
	and Margarita Chli$\,^{1}$
	\thanks{$^{1}\,$Authors are with the Vision for Robotics Lab (V4RL), University of Cyprus, Cyprus and ETH Zurich, Switzerland {\tt\footnotesize \{border.rowan,margarita.chli\}@ucy.ac.cy}}
}

\begin{document}
	
	\newacronym{lio}{LIO}{LiDAR Inertial Odometry}
	\newacronym{icp}{ICP}{Iterative Closest Point}
	\newacronym{imu}{IMU}{Inertial Measurement Unit}
	\newacronym{iekf}{iEKF}{iterated Extended Kalman Filter}
	\newacronym{tv}{TV}{Tensor Voting}
	\newacronym{ncd}{NCD}{Newer College Dataset}
	\newacronym{ape}{APE}{Absolute Pose Error}
	\newacronym{rmse}{RMSE}{Root Mean Square Error}
	
	\entryneedsurl{grupp2017}
	
	\maketitle
	
	\IEEEpeerreviewmaketitle
	
	\begin{abstract}
		\gls{lio} is a critical component for many mobile robots that need to navigate without relying on external positioning (e.g., GPS). Platforms that operate autonomously in different environments and with heterogeneous LiDAR sensors require a \gls{lio} approach that can adapt to these different scenarios without human intervention. 
		
		Existing \gls{lio} approaches can typically provide reliable and accurate odometry in scenarios with similar environments and sensors when suitably tuned. However, many approaches struggle to retain robust odometry across heterogeneous environments and sensors while using a consistent configuration. 
		
		This paper presents EllipseLIO, a real-time \gls{lio} approach that generalises between scenarios by using methods for LiDAR scan filtering and registration that adapt to the sensor capabilities and environment without requiring scenario-specific tuning. Experiments with EllipseLIO and state-of-the-art \gls{lio} approaches on five datasets with diverse and challenging scenarios demonstrate that EllipseLIO is the best performing approach overall. It achieves a $35\%$ lower odometry error on average than the second-best approach and is the only approach that does not diverge in any experiment. An open-source version of EllipseLIO is available at \url{https://github.com/v4rl-ucy/ellipselio}.
	\end{abstract}

	
	\glsresetall
	
	\section{Introduction}
	
	\IEEEPARstart{M}{obile} robots require a positioning system to safely navigate through their surrounding environment. Platforms that operate outdoors often use an external GNSS positioning system (e.g., GPS), but this is not possible when the signals are blocked by structures (e.g., indoors) or experience interference (i.e., jamming). \gls{lio} approaches enable robots to navigate using onboard odometry computed from LiDAR and \gls{imu} measurements. The reliability of the LiDAR odometry used is particularly critical for autonomous platforms that need to operate in diverse environments and with different sensors without human intervention.
	
	Existing \gls{lio} approaches can provide highly accurate and robust odometry in a wide range of scenarios, with different environments and sensors, when they are tuned to operate reliably in each scenario. However, many approaches do not generalise well between different scenarios without tuning as their methods for scan filtering and registration do not adapt online to the sensor capabilities and environment structure. This is substantiated by the variation in relative performance of the compared \gls{lio} approaches on different datasets \figref{marquee}. Scan filtering methods typically downsample LiDAR scans to a fixed resolution without considering the sensor resolution or the range of measurements. This can produce sparse scans with insufficient geometric fidelity for accurate registration or dense scans that are too computationally expensive to process online. Most scan registration methods use fixed error metrics that do not account for geometric variations and make planar surface assumptions, which can degrade odometry accuracy in environments where these assumptions are invalid. This is evidenced by the decrease in relative performance of DLIO~\cite{Chen2022}, FAST-LIO2~\cite{Xu2021}, and iG-LIO~\cite{Chen2024}, which make planar surface assumptions, between the structured Oxford Spires~\cite{Tao2025} dataset and the unstructured Botanic Garden~\cite{Liu2024b} dataset \figref{marquee}.    
	
	\begin{figure}[tpb]
		\centering
		\includegraphics[width=0.9\linewidth]{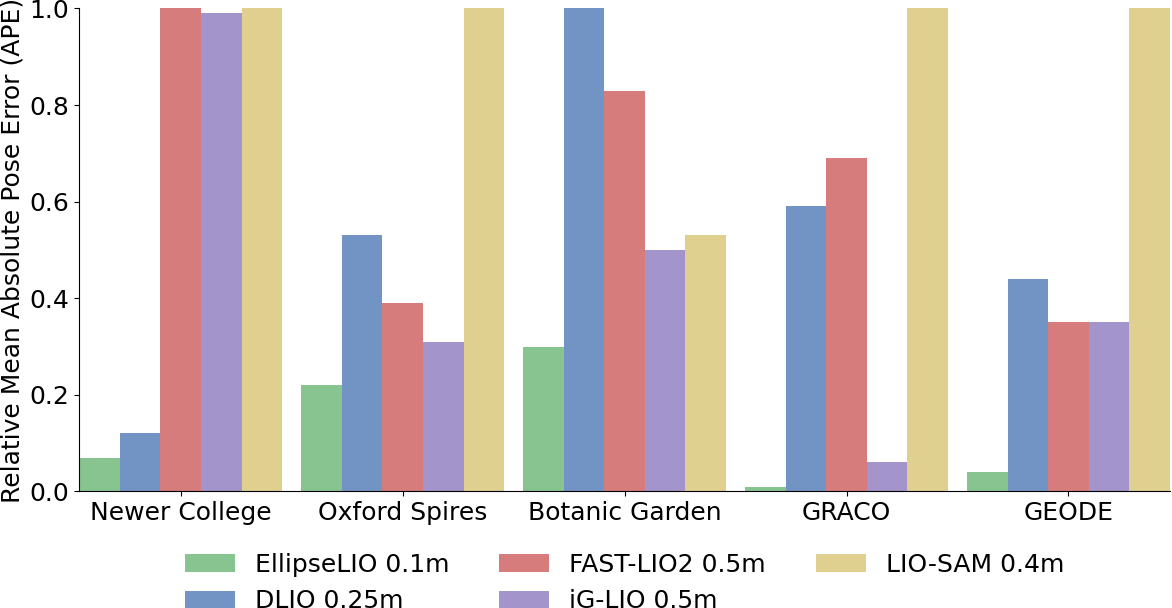}
		\caption{Results for EllipseLIO and the compared \gls{lio} approaches on five datasets with heterogeneous scenarios. The Relative Mean \gls{ape} is defined as the mean \gls{ape} of an approach on a dataset divided by the greatest mean \gls{ape} from all of the approaches, such that the values are scaled to $(0,1]$. EllipseLIO is consistently the best-performing approach, while the relative performance of the compared approaches varies between the datasets.}
		\figlabel{marquee}
		\vspace{-2ex}
	\end{figure}
	
	The approach presented in this paper, \textit{EllipseLIO}, addresses these challenges by using adaptive methods for scan filtering and registration that generalise between different scenarios. It makes the following key contributions: 
	\begin{itemize}
		\item An adaptive scan filtering method that retains sufficient measurements from a sensor to enable accurate registration while preserving computational efficiency by adjusting the downsampling resolution used at different ranges based on the sensor resolution and field-of-view.
		\item An ellipsoid-based scan registration method that provides reliable odometry estimates across diverse environments by adapting the error metric used for each point match to account for the local surface geometry.
		\item A drift correction method that is directly integrated into the scan registration pipeline and does not require separate loop closure handling. Point matches are adaptively weighted to prioritise those with older ellipsoids in revisited locations and correct accumulated drift. 
	\end{itemize}
	
	The performance of EllipseLIO is compared with existing state-of-the-art \gls{lio} approaches on five datasets with different environments, platforms, and sensors. The results demonstrate that EllipseLIO is able to provide more robust and accurate odometry estimates than the compared approaches across these diverse and challenging scenarios \figref{marquee}.
	
	\section{Related Work}
	
	\gls{lio} approaches aim to provide accurate and robust odometry by combining low-frequency but high-fidelity geometric information from LiDAR scans with high-frequency but noisy motion measurements from an \gls{imu}. They consist of three key stages: preprocessing, odometry, and mapping.
	
	The preprocessing stage applies filtering and in some cases motion-undistortion to the raw LiDAR scans. These scans can contain over a hundred thousand points and are too computationally expensive to process in real time. Some approaches~\cite{Zhang2014,Shan2020,Xu2020a} reduce this complexity by extracting geometric features (e.g., lines and planes), but these often do not generalise between different sensors. 
	
	Other approaches~\cite{Xu2021,Chen2022,Chen2024} downsample the points to a uniform resolution using a voxel grid filter. This is an efficient method for reducing the computation cost, but using a fixed resolution does not generalise well between different size environments. AdaLIO~\cite{Lim2023} presents a method for switching between resolutions in small- and large-scale environments, but it requires heuristic parameter tuning and still applies the same downsampling resolution to the entire scan. This paper presents a range-based filtering method that adaptively varies the downsampling resolution used at different ranges based on the LiDAR field-of-view and resolution.           
	
	The odometry stage estimates the sensor motion from the LiDAR scans and \gls{imu} measurements. In tightly coupled approaches, the \gls{imu} measurements propagate the odometry estimate between LiDAR scans. An estimate from the LiDAR is obtained by matching features or points between scans or against an incrementally updated map. Feature matching~\cite{Zhang2014,Shan2020,Xu2020a} can be computationally efficient, but its performance depends on the saliency of features extracted from a scan.
	
	Most approaches~\cite{Xu2021,Chen2022,Chen2024} perform scan registration using variants of the \gls{icp} algorithm. \gls{icp} associates points in the current scan with points in the previous scan or map and computes an alignment between them by minimizing the point-to-point, point-to-plane, or plane-to-plane distance. Point-to-point \gls{icp} works well in unstructured environments but takes longer to converge and is sensitive to noise. Point-to-plane \gls{icp} works well in structured environments and converges more quickly but is less accurate in unstructured environments. Generalized \gls{icp} (GICP)~\cite{Segal2009} uses a plane-to-plane error metric that is more robust to sensor noise in unstructured environments but still makes planar assumptions about the surface geometry.
	
	Recent work has focused on reducing drift and preventing divergence when using these \gls{icp} variants. X-ICP~\cite{Tuna2024} attempts to detect degenerate scenarios and prevent divergence by constraining the registration optimisation along degenerate axes. MAD-ICP~\cite{Ferrari2024} improves the robustness of point-to-plane \gls{icp} by directly encoding planar segments and surface normals into the map representation. GenZ-ICP~\cite{Lee2024} and Super Odometry~\cite{Zhao2021} combine point-to-point, point-to-plane, and point-to-line (only~\cite{Zhao2021}) metrics with weights derived from threshold parameters. The ellipsoid-based registration method presented in this paper adapts the error metric used for each point based on the shape of its associated ellipsoid.
	
	The mapping stage aggregates points from LiDAR scans into a map with a common reference frame that can be used for scan-to-map registration. The map is stored in a data structure that supports efficient searching and point insertion (e.g., an ikd-tree~\cite{Xu2020a,Xu2021}, Octree~\cite{Nguyen2022} or voxel hash map~\cite{Chen2024}). Many approaches use this data structure to directly represent their map but some also build a higher-level representation to encode additional geometric information (e.g., with surfels~\cite{Ramezani2022,Nguyen2022}). EllipseLIO stores map points in a dynamic Octree data structure~\cite{Zhu2024} and builds an ellipsoid map representation to encode high-fidelity geometric information.  
	
	\section{EllipseLIO}
	\seclabel{overview}
	
	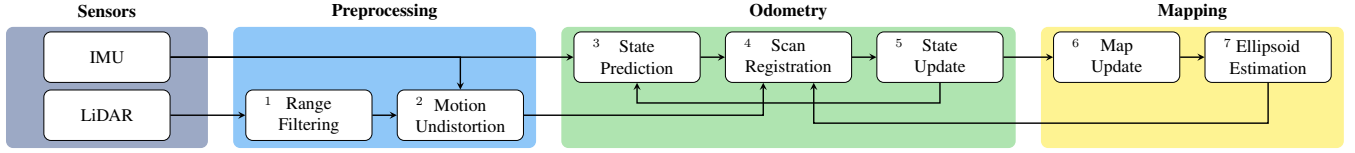
\begin{figure*}[tpb]
		\centering
		\resizebox{1.0\textwidth}{!}{%
		\begin{tikzpicture}[node distance=2.5cm, auto, >=latex]
			
			\node (sensors_bg) [stage, fill=viridis25!50, minimum width=4cm, minimum height=2.4cm] at (0, 0) {};
			\node (preprocessing_bg) [stage, fill=viridis50!50, minimum width=6cm, minimum height=2.4cm, right of=sensors_bg, xshift=3cm] {};
			\node (odometry_bg) [stage, fill=viridis75!50, minimum width=9cm, minimum height=2.4cm, right of=preprocessing_bg, xshift=5.5cm] {};
			\node (mapping_bg) [stage, fill=viridis100!50, minimum width=6cm, minimum height=2.4cm, right of=odometry_bg, xshift=5.5cm] {};
			
			\node at ([yshift=0.3cm]sensors_bg.north) {\textbf{Sensors}};
			\node at ([yshift=0.3cm]preprocessing_bg.north) {\textbf{Preprocessing}};
			\node at ([yshift=0.3cm]odometry_bg.north) {\textbf{Odometry}};
			\node at ([yshift=0.3cm]mapping_bg.north) {\textbf{Mapping}};
			
			\node (lidar) [process] at ([xshift=0cm,yshift=-0.55cm]sensors_bg.center) {LiDAR};
			\node (imu) [process, above of=lidar, xshift=0cm,yshift=-1.35cm] {\gls{imu}};
			
			\node (filtering) [process, label={[xshift=-0.8cm,yshift=-0.55cm]$^1$}] at ([xshift=-1.5cm,yshift=-0.55cm]preprocessing_bg.center) {Range Filtering};
			\node (undistortion) [process, label={[xshift=-0.8cm,yshift=-0.55cm]$^2$}, right of=filtering, xshift=0.5cm,yshift=-0cm] {Motion Undistortion};
			
			\node (prediction) [process, label={[xshift=-0.8cm,yshift=-0.55cm]$^3$}] at ([xshift=-3cm,yshift=0.6cm]odometry_bg.center) {State Prediction};
			\node (registration) [process, label={[xshift=-0.8cm,yshift=-0.55cm]$^4$}, right of=prediction, xshift=0.5cm] {Scan Registration};
			\node (update) [process, label={[xshift=-0.8cm,yshift=-0.55cm]$^5$}, right of=registration, xshift=0.5cm] {State\\Update};
			
			\node (mapupdate) [process, label={[xshift=-0.8cm,yshift=-0.55cm]$^6$}] at ([xshift=-1.5cm,yshift=0.6cm]mapping_bg.center) {Map\\Update};
			\node (geometry) [process, label={[xshift=-0.8cm,yshift=-0.55cm]$^7$}, right of=mapupdate, xshift=0.5cm] {Ellipsoid Estimation};
			
			\draw [arrow] (lidar) -- (filtering);
			\draw [arrow] (imu) -- (prediction);
			\draw [arrow] (imu.east) -- ++(0,0) -| (undistortion.north);
			\draw [arrow] (filtering) -- (undistortion);
			\draw [arrow] (undistortion.east) -- ++(0,0) -| ([xshift=-0.5cm]registration.south);
			\draw [arrow] (prediction) -- (registration);
			\draw [arrow] (registration) -- (update);
			\draw [arrow] (update) -- (mapupdate);
			\draw [arrow] (mapupdate) -- (geometry);
			
			\draw [arrow] (update.south) -- ++(0,-0.4) -| (prediction.south);
			\draw [arrow] (geometry.south) -- ++(0,-0.8) -| ([xshift=0.5cm]registration.south);
			
		\end{tikzpicture}
		}
		\caption{The EllipseLIO pipeline: (1) LiDAR scans are processed with range-based filtering and sent for motion undistortion, and (2) the processed scans are motion undistorted using predicted states. (3) Predicted states are forward-propagated from the last updated state using \gls{imu} measurements, while (4) processed scans are registered with the map to obtain a new pose estimate, and (5) the odometry state is updated using the pose estimate from the scan registration. (6) The map is updated with new scan points, while (7) new ellipsoids are initialised and existing ellipsoids are updated.}
		\figlabel{ellipselio}
		\vspace{-2ex}
	\end{figure*}
	
	EllipseLIO aims to provide reliable odometry across diverse scenarios with different environments and sensors by adapting to the environment and sensor capabilities. It applies adaptive range-based filtering to LiDAR scans instead of uniform resolution filtering to preserve sufficiently high-fidelity geometric information on surfaces at every range. It adapts the error metric used for each point match during the scan-to-map registration based on the structure of an ellipsoid that represents the local surface geometry around the map point. Accumulated drift is corrected when the sensor returns to a previously visited location by adaptively weighting point matches to prioritise those with older map points that were captured when the location was previously observed. An overview of the EllipseLIO pipeline is presented in Figure~\ref{fig:ellipselio}.                   
	
	\subsection{Preprocessing}
	\seclabel{scan-proc}
	
	\begin{figure}[tpb]
		\centering
		\begin{tikzpicture}[scale=1]
			\footnotesize
			
			\filldraw[black] (0,0) circle (1pt) node[above, yshift=0.1cm] {LiDAR};
			
			\draw[thick, black, line width=0.5mm] (-3, 2) -- (-3, -0.5) node[midway, above] {};
			\draw[thick, black, line width=0.5mm] (3, 2) -- (3, -0.5) node[midway, above] {};
			
			\draw[thick, black, line width=0.5mm] (-3, 2) -- (3, 2) node[midway, above] {};
			\draw[thick, black, line width=0.5mm] (-3, -0.5) -- (3, -0.5) node[midway, above] {};
			
			\draw [thick,<->,>=stealth] (-2,0) -- node[above] {$1\,$m} (-1,0);
			
			\foreach \x in {-2.75, -2.25, -1.7, -1.4, -1.1, -0.7, -0.5, -0.3, -0.1, 0.1, 0.3, 0.5, 0.7, 1.1, 1.4, 1.7, 2.25, 2.75} {
				\filldraw[gray] (\x, -0.5) circle (2pt);
			}
			\foreach \x in {-3, -2, -1.25, -0.5, 0, 0.5, 1.25, 2, 3} {
				\filldraw[gray] (\x, 2) circle (2pt);
			}
			\foreach \y in {0.5, 0, 0.5, 1.25, 2} {
				\filldraw[gray] (-3, \y) circle (2pt);
			}
			\foreach \y in {0.5, 0, 0.5, 1.25, 2} {
				\filldraw[gray] (3, \y) circle (2pt);
			}
			
			\clip (-3,-0.5) rectangle (3,2);
			\draw[gray] (0,0) circle (1);
			\draw[gray] (0,0) circle (2);
			\draw[gray] (0,0) circle (3);
			\draw[gray] (0,0) circle (4);
			
		\end{tikzpicture}
		\caption{An illustration of the range-based filtering approach, where the thick black lines represent surfaces and the grey dots are LiDAR points. It shows how the downsampling resolution is varied between each radial bin in order to retain more points on surfaces closer to the LiDAR.}
		\figlabel{range_filter}
		\vspace{-2ex}
	\end{figure}
	
	LiDAR scans typically contain over a hundred thousand points and need to be downsampled in order to provide real-time odometry estimates. These scans capture greater point density from nearby surfaces, which provides high-fidelity geometric information for small surface regions, and lower point density on distant surfaces, which can still provide valuable geometric information for larger surface regions. Applying uniform resolution downsampling to a scan removes more points from nearby surfaces, which significantly reduces the geometric fidelity, and often retains more distant points than are necessary to represent larger surface regions.     
	
	EllipseLIO addresses the limitations of uniform resolution downsampling by introducing a range-based method for filtering LiDAR scans that varies the downsampling resolution applied to points with their distance from the sensor. This method retains more points on nearby surfaces to represent small surface regions with high fidelity and fewer points on distant surfaces to represent larger regions with lower fidelity. It enables real-time processing by reducing the number of scan points while preserving enough geometric information from surfaces at every distance to achieve accurate scan registration in environments with different scales.
	
	The range-based filter segments a LiDAR pointcloud $P$ into $1\,$m wide radial bins \figref{range_filter}, such that the pointcloud for the $i$-th bin, $P_i$, contains the points between $i$ and $i+1$ meters from the sensor, $P_i = \{\mathbf{p} \in P\;|\;i < ||\mathbf{p}|| \le i+1 \}$. Downsampling is performed by inserting each pointcloud $P_i$ into an iOctree~\cite{Zhu2024} to obtain a filtered pointcloud $P^\prime_i$. The iOctree for the $i$-th bin is initialised with a voxel resolution $v_i$ and retains one point per voxel. No voxel-wise averaging of points is applied in order to preserve the original positions and measurement timestamps of the filtered points. The downsampling resolution $v_i$ for the $i$-th bin is given by the arc length between the scan lines of a repeating pattern LiDAR when they lie on the surface of a sphere with a radius of $i+1$~m, 
	\begin{equation}v_i = \frac{(i+1)\theta}{\beta - 1} \;,
		\eqlabel{res}
	\end{equation} 
	where $\theta$ is the vertical field-of-view of the LiDAR and $\beta$ is the number of scan lines. For a non-repeating LiDAR, $\beta$ is set to the same value as a repeating LiDAR with a similar point density. A suitable value can be obtained from manufacturer-provided metrics on field-of-view coverage. 
	
	This downsampling method produces a more uniform local point distribution within each bin, particularly for repeating LiDARs, by increasing the distance between points in the horizontal plane of the LiDAR to match the distance between points in the vertical plane. The filtered pointclouds from each bin, $P^\prime_i$, are aggregated into a combined pointcloud, $P^\prime$, and undistorted to compensate for sensor motion by using the method presented in FAST-LIO~\cite{Xu2020a}.                 
	
	\subsection{Odometry}
	\seclabel{odom}
	
	EllipseLIO performs state estimation using the tightly coupled \gls{iekf} method presented in FAST-LIO~\cite{Xu2020a}. The state $\mathbf{x}$ of the \gls{imu} frame, $I$, in the global frame, $G$, is  
	\begin{equation}\mathbf{x} = \left[\prescript{G}{}{\boldsymbol{\tau}}_{I}, \prescript{G}{}{\mathbf{R}}_{I}, \prescript{G}{}{\boldsymbol{\nu}}_{I}, \boldsymbol{\sigma}^{a}, \boldsymbol{\sigma}^{g}, \boldsymbol{\gamma}\right] \;,\end{equation}
	where $\boldsymbol{\tau} \in \mathbb{R}^3$ is the translation, $\mathbf{R} \in \mathit{SO}(3)$ is the rotation, $\boldsymbol{\nu} \in \mathbb{R}^{3}$ is the linear velocity, $\boldsymbol{\sigma}^{a}, \boldsymbol{\sigma}^{g} \in \mathbb{R}^3$ are the accelerometer and gyroscope biases, and $\boldsymbol{\gamma} \in \mathbb{R}^{3}$ is the gravity vector. Updated states are denoted as $\bar{\mathbf{x}}_i$ and are obtained by registering a new LiDAR scan with the map. Predicted states are denoted as $\widehat{\mathbf{x}}_{i+n}$ and are obtained by propagating the previous updated state $\bar{\mathbf{x}}_{i}$ forwards using new \gls{imu} measurements.
	
	 The scan-to-map registration computes a transformation of the scan that minimizes its alignment error with the map. Points from the combined pointcloud, $\prescript{L}{}{\mathbf{p}} \in P^\prime$, are transformed into the global frame from the LiDAR frame $L$, using the latest predicted state $\widehat{\mathbf{x}}_k$,
	\begin{equation}\prescript{G}{}{\mathbf{p}} = \prescript{G}{}{\mathbf{T}}^k_{I}\prescript{I}{}{\mathbf{T}}_{L}\prescript{L}{}{\mathbf{p}}\;,
		\eqlabel{tf} 
	\end{equation}
	where $\prescript{T}{}{\mathbf{T}}^i_{S} = [\prescript{T}{}{\mathbf{R}}^i_{S} \,\ | \, \prescript{T}{}{\boldsymbol{\tau}}^i_{S} ]$ is the transformation from a source frame $S$, to a target frame $T$, at time $t_i$. 
	
	Each scan point in the global frame, $\prescript{G}{}{\mathbf{p}}$, is matched with its closest map point, $\prescript{G}{}{\mathbf{q}}$, by performing a nearest-neighbour search within a radius $r$. This radius is defined by the radial bin of the scan point (Alg. 1, Line 4; \tsecref{map}) and is incrementally reduced for each iEKF iteration.
	
	The error metric used for each match is determined by the structure of an ellipsoid associated with the map point. This ellipsoid represents the local surface geometry around the map point and is estimated from neighbouring points in the mapping stage \secref{map}. Map points with insufficient neighbours have no associated ellipsoid and newly created ellipsoids may not yet accurately represent the local geometry, so these matches are excluded from the registration.
	
	\begin{figure}[tpb]
		\centering
		\includegraphics[width=0.82\linewidth]{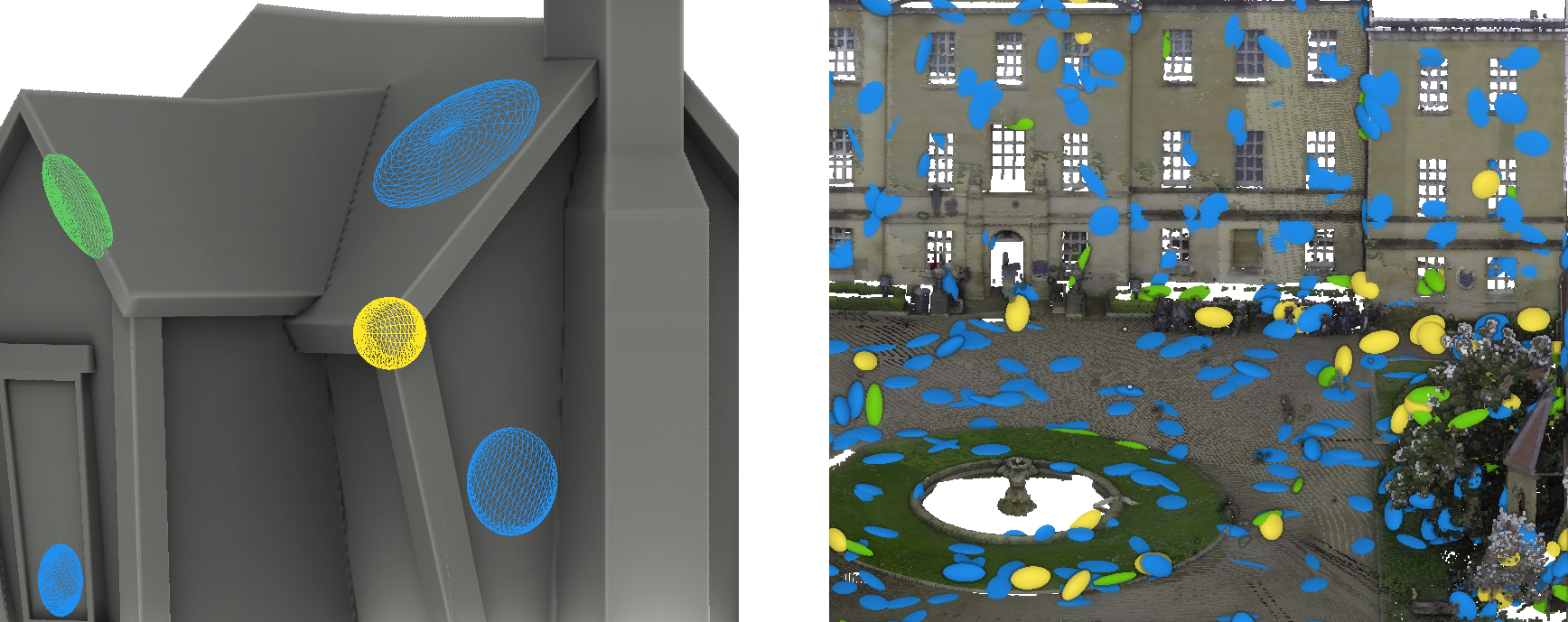}
		\caption{Illustrated (left) and real (right) ellipsoids classified by their most salient geometric primitive: \emph{lines} (green), \emph{planes} (blue), and \emph{balls} (yellow). This shows how the ellipsoids represent the underlying surface geometry.}
		\figlabel{geo_prim}
		\vspace{-2mm}
	\end{figure}  
	
	The ellipsoid structure is defined by three unit vectors $\mathbf{V} = \{\mathbf{v}_1, \mathbf{v}_2, \mathbf{v}_3\}$, and their corresponding magnitudes $\mathbf{m} = \{m_1, m_2, m_3\}$, where $m_1 \le m_2 \le m_3$. This structure is decomposed into geometric primitives that represent how closely the ellipsoid shape matches a \emph{line}, \emph{plane}, or \emph{ball} \figref{geo_prim}. Line primitives denote surfaces with a single dominant axis of variation (e.g., a pole), plane primitives denote surfaces with minimal variation along a single axis (i.e., the surface normal), and ball primitives denote surfaces with similar variation along each axis. These primitives are assigned saliency scores, $\mathbf{g} = [g_\mathrm{line}, g_\mathrm{plane}, g_\mathrm{ball}]$, which are used to weight registration metrics for each primitive.
	
	Line primitives use a point-to-line metric, which projects the scan point onto a line through the map point whose direction is given by the ellipsoid axis with the largest magnitude $\mathbf{v}_3$,
	\begin{equation}\prescript{G}{}{\mathbf{p}^\prime_\mathrm{line}} = \prescript{G}{}{\mathbf{q}} + ((\prescript{G}{}{\mathbf{p}} - \prescript{G}{}{\mathbf{q}}) \cdot \mathbf{v}_3) \mathbf{v}_3 \;.
	\end{equation}
	Plane primitives use a point-to-plane metric, which projects the scan point onto a plane whose normal vector is given by the ellipsoid axis with the smallest magnitude, $\mathbf{v}_1$,  
	\begin{equation}\prescript{G}{}{\mathbf{p}^\prime_\mathrm{plane}} = \prescript{G}{}{\mathbf{p}} - ((\prescript{G}{}{\mathbf{p}} - \prescript{G}{}{\mathbf{q}}) \cdot \mathbf{v}_1) \mathbf{v}_1 \;.
	\end{equation}
	Ball primitives use a point-to-point metric, which projects the scan point directly onto its matched map point,
	\begin{equation}\prescript{G}{}{\mathbf{p}^\prime_\mathrm{ball}} = \prescript{G}{}{\mathbf{q}} \;.
	\end{equation}
	The resulting target point $\prescript{G}{}{\mathbf{p}}'$, is the sum of the projected scan points for each geometric primitive, weighted by their L1-normalised saliency scores,
	\begin{equation}\prescript{G}{}{\mathbf{p}^\prime} = \frac{g_\mathrm{line}}{||\mathbf{g}||_1}\prescript{G}{}{\mathbf{p}^\prime_\mathrm{line}} + \frac{g_\mathrm{plane}}{||\mathbf{g}||_1}\prescript{G}{}{\mathbf{p}^\prime_\mathrm{plane}} + \frac{g_\mathrm{ball}}{||\mathbf{g}||_1}\prescript{G}{}{\mathbf{p}^\prime_\mathrm{ball}} \;.
	\end{equation} 
	
	The \gls{iekf} state is updated by populating a residual vector $\mathbf{z} = \left[z_0, \dots, z_n\right]$, a measurement matrix $\mathbf{H} = \left[\mathbf{h}_0, \dots, \mathbf{h}_n\right]$, and a weight vector $\mathbf{w} = \left[w_0, \dots, w_n\right]$, similar to~\cite{Xu2020a}.
	
	The residual error for a match is given by the Euclidean distance between the scan point and its target point,
	\begin{equation}z_i = -\left|\left|\prescript{G}{}{\mathbf{p}}_i - \prescript{G}{}{\mathbf{p}}'_i\right|\right| \;.
	\end{equation}
	
	The measurement vector for a match is structured as $\mathbf{h}_i = \left[\mathbf{t}_i^\top\;\mathbf{r}_i^\top\right]$, where $\mathbf{t}_i \in \mathbb{R}^3$ defines the translation and $\mathbf{r}_i \in \mathit{SO}(3)$ defines the rotation.
	
	The translation component is given by a unit vector from the target point to the scan point,
	\begin{equation}\mathbf{t}_i = \frac{\prescript{G}{}{\mathbf{p}}_i - \prescript{G}{}{\mathbf{p}}'_i}{\left|\left|\prescript{G}{}{\mathbf{p}}_i - \prescript{G}{}{\mathbf{p}}'_i\right|\right|}\;.
	\end{equation}
	
	The rotation component is given by a cross product of the scan point with the translation component in the \gls{imu} frame,
	\begin{equation}\mathbf{r}_i = \prescript{I}{}{\mathbf{T}}^k_{G}\prescript{G}{}{\mathbf{p}}_i \times \prescript{I}{}{\mathbf{R}}^k_{G}\,\mathbf{t}_i\;.
	\end{equation}   

	\subsubsection{Adaptive Match Weighting}

	Each point match is weighted based on the distance traveled since the map point was captured. Map points with large travel distances indicate that the sensor has traveled farther and accumulated more odometry drift before returning to a location where some surfaces were previously observed. Matches with these map points are assigned greater weight to correct this drift.        
	
	The match weight $w_i$ is given by the difference in trajectory length $\mathrm{s}(t)$ between the measurement time of the scan point $t_{\mathbf{p}_i}$, and its matched map point $t_{\mathbf{q}}$. Matches with residual error in the vertical axis, where drift often accumulates faster as there are typically fewer constraints, are prioritised and given greater weight by scaling the match weights based on the dot product between the translational component of the match, $\widehat{\mathbf{t}}_i\,$, and the unit gravity vector, $\widehat{\boldsymbol{\gamma}}\,$,
	\begin{equation}w_i = \frac{\mathrm{s}(t_{\mathbf{p}_i}) - \mathrm{s}(t_{\mathbf{q}}) + 1}{\max\left(1 - | \widehat{\mathbf{t}}_i \cdot \widehat{\boldsymbol{\gamma}} |, 10^{-4}\right)}\;.
	\end{equation}

	Large match weights can increase the risk of divergence in geometrically degenerate environments where some state dimensions are poorly constrained. EllipseLIO adapts the degeneracy detection method presented in SuperLoc~\cite{Zhao2025} to estimate the translation and rotation state observability, $\mathbf{o}_i^\mathrm{t}$ and $\mathbf{o}_i^\mathrm{r}$, provided by the constraints of each match,    
	\begin{equation}\mathbf{o}_i^\mathrm{t} = \left(\frac{g_\mathrm{plane}}{||\mathbf{g}||_1}\right)^2 \,  \left(\frac{\mathbf{c}^\top}{\max(\mathbf{c})} \odot |\mathbf{t}_i^\top \prescript{\Gamma}{}{\mathbf{R}}_{L}|\right) \;,
	\end{equation}
	and
	\begin{equation}\mathbf{o}_i^\mathrm{r} = |(\prescript{G}{}{\mathbf{R}}^k_{I} \,\hat{\mathbf{r}}_i)^\top \prescript{\Gamma}{}{\mathbf{R}}_{L}|  \;,
	\end{equation}
	where $\mathbf{c} = \mathrm{diag}(\boldsymbol{\Sigma}_{P^{\prime\prime}})$ is the diagonal of the covariance matrix for the processed scan in the gravity-aligned frame $\Gamma$, $P^{\prime\prime} = \prescript{\Gamma}{}{P^\prime}$,  $\prescript{\Gamma}{}{\mathbf{R}}_{L}$ is the rotation from the LiDAR frame to the gravity-aligned frame and $\odot$ is the element-wise Hadamard product. Observability matrices, $\mathbf{O}^x = \left[\mathbf{o}_0^x, \dots, \mathbf{o}_n^x\right]$, are populated with the maximum translation and rotation observability values for each match,      
	\begin{equation}\mathbf{o}_i^x \gets \mathbf{o}_i^x \odot [\mathbf{o}_i^x = \max(\mathbf{o}_i^x)], \; x \in \{\mathrm{t}, \mathrm{r}\} \;.
	\end{equation}
	The observability value for each state dimension is the sum of the values for all matches, scaled by the maximum observability value for the translation or rotation states, 
	\begin{equation}\mathbf{o}^x = \frac{\mathbf{1}^\top \mathbf{O}^x}{\max\left(\mathbf{1}^\top \mathbf{O}^x\right)}, \; x \in \{\mathrm{t}, \mathrm{r}\}\;. 
	\end{equation}

	The match weights $\mathbf{w} = \left[w_0, \dots, w_n\right]$ are scaled to mitigate the risk of divergence caused by degeneracy and incorrect match associations, particularly in enclosed environments and those with changing elevation. The weights are first scaled to lie within one standard deviation of their mean and shifted to have a minimum value of $1$. Exponential penalties for degeneracy, $\eta = \min(\mathbf{o}^\mathrm{t})\min(\mathbf{o}^\mathrm{r})$, vertical velocity, $\psi = 1 - \min(| \boldsymbol{\nu} \cdot \widehat{\boldsymbol{\gamma}} |, 1)$, and the mean scan range, $\mu = \min(0.1 || \prescript{L}{}{\mathbf{p}_\mathrm{mean}} ||, 1)$, are then applied such that $\mathbf{w} \gets \mathbf{w}^{\circ (\eta \psi \mu)}$, where $\circ$ is the Hadamard power.
	
	The scan registration and \gls{iekf} update are iterated until the state change converges below a given threshold or the registration time exceeds half of the LiDAR scan duration. The new updated state $\bar{\mathbf{x}}_k$ is then returned.   
	
	\subsection{Mapping}
	\seclabel{map}
	
	Mapping aggregates the LiDAR scans into a combined map in the global frame using the updated odometry states. When points from a new scan are added to the map they are processed to estimate their associated ellipsoids and update the ellipsoids of neighbouring points. These ellipsoids need to represent the local surface geometry with sufficient fidelity to enable accurate scan registration at different ranges. This is achieved by varying the density of points added to the map and the size of the ellipsoids created based on their distance from the sensor, such that the ellipsoids for nearby points represent the geometry of a smaller surface area.        
	
	\subsubsection{Map Update}
	\seclabel{mapupdate}
	The map pointcloud, $Q := \{\mathbf{q}_i\}^n_{i=1}$, is stored in an iOctree~\cite{Zhu2024} with a fixed voxel resolution, $\phi = 0.1$, which provides a suitable upper bound for the memory usage of the map. Points from a new scan are added to the map after performing scan-to-map registration and obtaining an updated odometry state $\bar{\mathbf{x}}_k$, which is used to transform the LiDAR points into the global frame \eqrefb{tf}.        
	
	The maximum density of points added to the map within each radial bin is set to provide an upper bound on the memory usage while ensuring that each voxel can contain sufficient points to estimate an ellipsoid of the correct size at the corresponding radial distance. The maximum number of points per voxel $\rho_i$, for a bin $i$, is
	\begin{equation}
	\rho_i = \min\left\{\left\lceil n_\mathrm{max}\phi^2 (\pi r_i^2)^{-1}\right\rceil, n_\mathrm{min}\right\}\;,
	\end{equation}
	where $r_i$ is the search radius used to estimate ellipsoids within the bin (Alg. 1, Line 4; \tsecref{map}), $n_\mathrm{min} = 6$ is a fixed lower bound on the minimum number of points required to estimate an ellipsoid to ensure stability, and $n_\mathrm{max} = 60$ is a fixed upper bound on the maximum number of points that can be part of an ellipsoid to constrain the computation cost.
	
	The existing map $Q_{-1}$ is iteratively updated by inserting each bin pointcloud in the global frame, $\prescript{G}{}{P}'_i$, into the iOctree. For each bin, the iOctree returns an updated map, $Q_i$, and the set of newly added points, $Q^\prime_i$,
	\begin{equation}
	Q_i, Q^\prime_i \gets \mathrm{iOctree}\left(Q_{i-1}, \prescript{G}{}{P}'_i, \phi, \rho_i\right)\;,
	\end{equation}
	where $Q^\prime_i = Q_i \setminus Q_{i-1} $. This update is repeated for every bin, $0 \le i \le i_\mathrm{max}$, to obtain a final updated map, $Q = Q_{i_\mathrm{max}}$, and a combined set of new points, $Q^\prime$, by aggregating the newly added points from each bin, $Q^\prime_i$.
	
	\subsubsection{Ellipsoid and Primitive Estimation}
	
	EllipseLIO performs surface geometry estimation using the \gls{tv} method presented in~\cite{Medioni2000}. \gls{tv} estimates the surface geometry for a set of neighbouring points in two stages. 
	
	The first stage computes an initial estimate of the surface normal at each point based on the relative distances and positions of its neighbouring points. The second stage propagates Eigen decompositions of these initial estimates between the points as tensor votes, such that the final estimate of the surface geometry around each point is a weighted sum of the 2D tensors from its neighbouring points.                  
	
	The tensors are computed using the formulation presented in~\cite{Wu2016,Labussiere2018a}. This computes the initial $k$-th stage tensor $\mathbf{J}^k_i$, for a target point $\mathbf{q}_i$, as a weighted sum of the final $(k-1)$-th stage tensors $\mathbf{K}^{k-1}_{j}$, for its neighbouring points $\mathbf{q}_j \in \mathrm{N}(\mathbf{q}_i, Q, r)$, within a search radius $r$,   
	\begin{equation}
		\mathbf{J}^k_i = \sum_{\mathbf{q}_j \in \mathrm{N}(\mathbf{q}_i, Q, r)} c_{ij}\mathbf{U}_{ij}\mathbf{K}^{k-1}_{j}\mathbf{U}_{ij}^{\prime}\;,
		\eqlabel{tv}
	\end{equation}
	where $\mathrm{N}(\mathbf{q}_i, Q, r) := \{\mathbf{q}_j \in Q \,|\; ||\mathbf{q}_i - \mathbf{q}_j|| \le r \}$, $c_{ij} = \exp\left(-{||\mathbf{q}_i-\mathbf{q}_j||}^2 r^{-1} \right)$ is a scalar weight defined by the distance to the neighbouring point and the search radius, and $\mathbf{U}_{ij}$, $\mathbf{U}_{ij}^{\prime}$ account for the relative position of the point,  
	\begin{equation}
		\mathbf{U}_{ij} = \left(\mathbf{I}-2\widehat{\mathbf{u}}_{ij}\widehat{\mathbf{u}}_{ij}^\top\right),
		\mathbf{U}_{ij}^{\prime} = \left(\mathbf{I}-0.5\widehat{\mathbf{u}}_{ij}\widehat{\mathbf{u}}_{ij}^\top\right)\mathbf{U}^\top_{ij}\;,
	\end{equation}
	where $\widehat{\mathbf{u}}_{ij}$ is the unit vector for $\mathbf{u}_{ij} = \mathbf{q}_i-\mathbf{q}_j$.
	
	The zeroth stage tensor is initialized as $\mathbf{K}^{0}_i = \mathbf{I}_3$. The initial first stage tensor $\mathbf{J}^{1}_i$ is post-processed by applying an Eigen decomposition to produce a set of eigenvalues $\boldsymbol{\lambda} = \{\lambda_{1},\lambda_{2},\lambda_{3}\}$ and their corresponding eigenvectors $\boldsymbol{\Upsilon} = \{\boldsymbol{\upsilon}_{1}, \boldsymbol{\upsilon}_{2}, \boldsymbol{\upsilon}_{3}\}$, where $\lambda_1 \ge \lambda_2 \ge \lambda_3$. The final first stage tensor $\mathbf{K}^{1}_i$, is then obtained by extracting the component votes for the plane and line primitives, 
	\begin{equation}
		\mathbf{K}^{1}_i = \left(\lambda_1 - \lambda_2 \right) \boldsymbol{\upsilon}_1\boldsymbol{\upsilon}^\top_1 + \left(\lambda_2 - \lambda_3 \right) \sum_{d=1}^{2}\boldsymbol{\upsilon}_d\boldsymbol{\upsilon}^\top_d \;.
		\eqlabel{pt}
	\end{equation}
	
	The initial second stage tensor $\mathbf{J}^{2}_i$, is Eigen decomposed to determine the saliency of the local surface estimate as \emph{line}, \emph{plane}, and \emph{ball} primitives. The saliency of each primitive is given by the tensor eigenvalues. The eigenvalue magnitudes correspond to how consistent the normals for neighbouring points are with the orientation of each eigenvector. This differs from the eigenvalues of a pointcloud covariance matrix, which correspond to the magnitude of surface variation.
	
	Linear surfaces have normals that consistently lie on a plane perpendicular to the primary linear axis and their saliency is given by $\lambda_2 - \lambda_3$. Planar surfaces have normals with a consistent orientation and their saliency is given by $\lambda_1 - \lambda_2$. Ball-like surfaces have no consistent normal orientation and their saliency is given by $\lambda_3$. The saliencies of the geometric primitives, $\mathbf{g}_i$, associated with the point $\mathbf{q}_i$, are computed from these eigenvalues,
	\begin{equation}
		\mathbf{g}_i = [\lambda_2 - \lambda_3, \lambda_1 - \lambda_2, \lambda_3]\;.
		\eqlabel{prim}
	\end{equation}
	
	An ellipsoid representing the surface geometry around the point is obtained by inverting the eigenvalues to denote the magnitude of surface variation $\mathbf{m}$, in the direction of their associated eigenvector, instead of the normal consistency, 
	\begin{equation}
		\mathbf{m} = \left\{r (\lambda_i\lambda^\prime)^{-1}\,\middle|\, \lambda_i \in \boldsymbol{\lambda} \right\}\;,
		\eqlabel{ellipse}
	\end{equation}
	where the magnitude is normalized by the sum of the inverted eigenvalues $\lambda^\prime = \lambda^{-1}_1 + \lambda^{-1}_2 + \lambda^{-1}_3$ and scaled by the search radius, $r$. The ellipsoid axes are the eigenvectors $\mathbf{V} = \boldsymbol{\Upsilon}$. 
	
	\subsubsection{Ellipsoid Update}
	
	\setlength{\textfloatsep}{2ex}
	\begin{algorithm}[tpb]
		\scriptsize
		\caption{Ellipsoid Update}
		\begin{algorithmic}[1]
			\State $N^\prime \gets \emptyset$
			\For{$\mathbf{q}_i \in Q^\prime$}
			\State $b_i \gets \mathbf{b}(\mathbf{q}_i)$
			\State $r_i \gets \min\{10\mathbf{v}(b_i), \,1\}$
			\State $N_i \gets \mathrm{N}(Q, \mathbf{q}_i, r_i)$
			\vspace{0.25ex}
			\If{$|N_i| \ge n^{b_i}_\mathrm{min}$} 
			\vspace{0.25ex}
			\State $\mathbf{J}^1_i \gets$ ComputeTensor$(\mathbf{q}_i, N_i, 1)$\Comment{\eqrefb{tv}}
			\vspace{0.25ex}
			\State $\mathbf{K}^1_i \gets$ ProcessTensor$(\mathbf{J}^1_i)$\Comment{\eqrefb{pt}}
			\vspace{0.25ex}
			\Else
			\State $Q^\prime \gets Q^\prime \setminus \{\mathbf{q}_i\}$
			\EndIf
			\vspace{-0.25ex}
			\For{$\mathbf{q}_j \in N_i$}
			\State $b_j \gets \mathbf{b}(\mathbf{q}_j)$
			\State $r_j \gets \min\{10\mathbf{v}(b_j), \,1\}$
			\If {$|| \mathbf{q}_i - \mathbf{q}_j || \le r_j$ \textbf{and} $|N_j| < n^{b_j}_\mathrm{max}$}
			\State $N_j \gets N_j \cup \{\mathbf{q}_i\}$
			\If{$|N_j| \ge n^{b_j}_\mathrm{min}$ \textbf{and} $\mathbf{q}_j \notin Q'$}
			\vspace{0.25ex} 
			\State $N^\prime \gets N^\prime \cup \{\mathbf{q}_j\}$
			\vspace{-0.25ex}
			\EndIf
			\EndIf
			\EndFor
			\EndFor
			\For{$\mathbf{q}_j \in N^\prime$}
			\vspace{0.25ex}
			\State $\mathbf{J}^1_j \gets$ ComputeTensor$(\mathbf{q}_j, N_j, 1)$\Comment{\eqrefb{tv}}
			\vspace{0.25ex}
			\State $\mathbf{K}^1_j \gets$ ProcessTensor$(\mathbf{J}^1_j)$\Comment{\eqrefb{pt}}
			\vspace{-0.25ex}
			\EndFor
			\For{$\mathbf{q}_k \in Q' \cup N^\prime$}
			\vspace{0.25ex}
			\State $\mathbf{J}^2_k \gets$ ComputeTensor$(\mathbf{q}_k, N_k, 2)$\Comment{\eqrefb{tv}}
			\vspace{0.25ex}
			\State $\mathbf{g}_k \gets$ GetPrimitiveSaliencies$(\mathbf{J}^2_k)$\Comment{\eqrefb{prim}}
			\vspace{0.25ex}
			\State $\mathbf{m}_k, \mathbf{V}_k \gets$ GetEllipsoid$(\mathbf{J}^2_k)$\Comment{\eqrefb{ellipse}}
			\vspace{0.25ex}
			\EndFor
			\vspace{-0.25ex}
		\end{algorithmic}
		\algmlabel{geo_prim}
	\end{algorithm}
	
	The ellipsoids associated with map points are updated by initializing ellipsoids for newly added points and reprocessing the ellipsoids for neighbouring points \algmref{geo_prim}. The neighbours of each new point $\mathbf{q}_i \in Q^\prime$, within a radius $r_i$, are found by searching the iOctree (Line 5). The search radius is scaled with the range of the point to estimate smaller ellipsoids for nearby surfaces. It is $10\times$ the voxel resolution, $\mathbf{v}(b_i)$, of the radial bin associated with the point, $b_i = \mathbf{b}(\mathbf{q}_i)$, up to a radius of $1\,$m (Lines 3--4).
	
	An ellipsoid is only estimated if the point has a minimum number of neighbours $n^{b_i}_\mathrm{min}$, to represent the local surface geometry. The minimum for a radial bin is computed from the mean number of neighbours for points in the bin,
	\begin{equation}
		n^{b_i}_\mathrm{min} = \max\left\{n_\mathrm{min}, \sum_{\mathbf{q}_j \in \mathrm{B}(\mathbf{q}_i)} \frac{|\mathrm{N}(Q, \mathbf{q}_j, r_j)|}{|\mathrm{B}(\mathbf{q}_i)|}\right\}\;,
	\end{equation}
	where $\mathrm{B}(\mathbf{q}_i) := \left\{\mathbf{q}_j \in Q \,\middle|\, \mathbf{b}(\mathbf{q}_i) \equiv \mathbf{b}(\mathbf{q}_j) \right\}$. A first stage tensor is computed for new points with sufficient neighbours and those without are removed from the set $Q^\prime$ (Lines 6--10).
	
	The neighbours of each new point $\mathbf{q}_j \in N_i$, are processed to update their neighbourhood set $N_j$ (Line 11). A new point is added to the set of its neighbour if it lies within the search radius of the neighbour and the set has less than a maximum number of points $n^{b_j}_\mathrm{max} = \min\{n_\mathrm{max}, 2n^{b_j}_\mathrm{min}\}$ (Lines 12--15). A neighbour is added to the set of existing points for reprocessing, $N^\prime$, if it has the minimum number of required neighbours and is not a newly added point (Lines 16--17).
	
	After all the newly added points are processed the set of existing points is processed to update their first stage tensors (Lines 18--20). The combined set of newly added and neighbouring existing points is then processed to compute second stage tensors, extract the primitive saliency values and obtain the representative ellipsoids (Lines 21--24).
	
	The computational complexity of the ellipsoid update is $O(|Q^\prime|(\log|Q| + |N|))$, where $\log|Q|$ represents the search complexity for the iOctree and $|N|$ is the number of neighbours for each new point (Line 5). The update process is efficient as all of the loops are fully parallelisable.                    
	
	\section{Evaluation}
	\seclabel{evaluation}
	
	EllipseLIO is evaluated on five datasets: Newer College~\cite{Ramezani2020, Zhang2021a}, Oxford Spires~\cite{Tao2025}, Botanic Garden~\cite{Liu2024b}, GRACO~\cite{Zhu2023}, and GEODE~\cite{Chen2025}. These cover a diverse range of environments and both spinning and solid-state LiDARs. The Newer College and Oxford Spires datasets capture mostly structured urban scenes with Ouster OS1-64 and OS0-128 LiDARs. The Botanic Garden dataset captures an unstructured park with a Velodyne VLP-16. The GRACO dataset captures an urban area from an aerial platform with a VLP-16. The GEODE dataset contains challenging sequences from diverse environments with spinning LiDARs, a VLP-16 (\emph{alpha}) and an OS1-64 (\emph{beta}), and a solid-state Livox Avia (\emph{gamma}) with a limited horizontal field-of-view.   
	
	The performance of EllipseLIO is compared with state-of-the-art \gls{lio} approaches: DLIO~\cite{Chen2022}, FAST-LIO2~\cite{Xu2021}, LIO-SAM \cite{Shan2020}, and iG-LIO~\cite{Chen2024}, using their open-source implementations. All of the approaches were run with ROS2 Humble and Ubuntu 22.04 on an Intel i9-13905H CPU with $32\,$GB RAM. Comparisons with LiDAR odometry approaches (e.g., MAD-ICP~\cite{Ferrari2024} and GenZ-ICP~\cite{Lee2024}) are not presented as they have a lower overall performance than the \gls{lio} approaches.
	
	EllipseLIO and the compared approaches use fixed parameters for all of the dataset sequences. EllipseLIO uses its predefined map resolution and a scan filtering resolution computed using the range-based adaptive method. The Livox Avia scan-line parameter is set to $\beta = 64$ by matching its manufacturer-provided field-of-view coverage to a spinning LiDAR with similar capabilities. The compared approaches use the voxel resolutions specified in Table 1 for both their map and scan filtering. Each compared approach is run with the same $0.1\,$m map resolution as EllipseLIO to provide a direct comparison and a tuned resolution that produces the best overall performance; DLIO uses $0.25\,$m, FAST-LIO2 uses $0.5\,$m, LIO-SAM uses $0.4\,$m, and iG-LIO uses $0.5\,$m.
	
	\glsreset{ape}
	The odometry performance of the approaches is compared by evaluating the \gls{ape} between their pose estimates and ground truth poses for the dataset sequences. The odometry and ground-truth trajectories were aligned with $\mathit{SE}(3)$ Umeyama alignment and the \gls{rmse} between corresponding odometry and ground truth poses was then computed using evo~\cite{grupp2017}.
	
	\section{Discussion}
	
	\begin{table}[tpb]
	\tbllabel{perf}
	\centering
	\scriptsize
	\adjustbox{width=\columnwidth,center}{%
		\addtolength{\tabcolsep}{-0.3em}
		\begin{tabular}{@{}clccccc@{}}
			&  & \multicolumn{5}{c}{Odometry Performance (APE RMSE in meters)} \\
			\toprule
			&  & EllipseLIO & DLIO & FAST-LIO2 & LIO-SAM & iG-LIO \\
			& Voxel Resolution (m) & 0.1 & 0.1 / 0.25 & 0.1 / 0.5 & 0.1 / 0.4 & 0.1 / 0.5 \\ \midrule
			\multirow{8}{*}{\rotatebox[origin=c]{90}{Newer College}} & short-experiment & \textbf{0.30} & 0.45 / 0.44 & $\;\,\times\;$ / 0.41 & $\;\,\times\;$ / 0.41 & 0.51 / \underline{0.34} \\
			& quad-with-dynamics & \underline{0.09} & 0.14 / 0.15 & $\;\,\times\;$ / 0.12 & \textbf{0.08} / \textbf{0.08} & \underline{0.09} / \underline{0.09} \\
			& dynamic-spinning & \textbf{0.08} & 0.15 / 0.15 & $\;\,\times\;$ / \textbf{0.08} & 0.10 / 0.10 & \textbf{0.08} / \underline{0.09} \\
			& parkland-mound & \textbf{0.12} & 0.19 / 0.19 & $\;\,\times\;$ / \textbf{0.12} & $\;\,\times\;$ / \underline{0.13} & $\;\,\times\;$ / 0.14 \\
			& stairs & \textbf{0.08} & \textbf{0.08} / \underline{0.10} & \textbf{0.08} / $\;\,\times\;$ & 2.68 / $\;\,\times\;$ & $\;\,\times\;$ / $\;\,\times\;$ \\
			& cloister & \textbf{0.07} & 0.21 / 0.20 & $\;\,\times\;$ / \underline{0.08} & $\;\,\times\;$ / 0.12 & 0.26 / 0.12 \\
			& quad-hard & \textbf{0.07} & 0.16 / 0.12 & $\;\,\times\;$ / 0.11 & 1.49 / 0.12 & $\;\,\times\;$ / \underline{0.10} \\ \midrule
			\multirow{5}{*}{\rotatebox[origin=c]{90}{\parbox{1cm}{\centering Oxford Spires}}} & blenheim-01 & \underline{0.14} & 0.46 / 0.31 & $\;\,\times\;$ / \underline{0.14} & $\;\,\times\;$ / 0.52 & $\;\,\times\;$ / \textbf{0.12} \\
			& bodleian-02 & \textbf{0.19} & 0.75 / 0.54 & $\;\,\times\;$ / 0.54 & 0.42 / 1.32 & $\;\,\times\;$ / \underline{0.32} \\
			& christchurch-03 & \textbf{0.03} & 0.08 / 0.07 & $\;\,\times\;$ / \underline{0.04} & $\;\,\times\;$ / 0.08 & $\;\,\times\;$ / 0.08 \\
			& keble-02 & \textbf{0.05} & 0.08 / \underline{0.06} & $\;\,\times\;$ / \textbf{0.05} & $\;\,\times\;$ / 0.13 & 0.13 / 0.07 \\
			& observatory-01 & \textbf{0.06} & 0.27 / 0.18 & $\;\,\times\;$ / \underline{0.08} & $\;\,\times\;$ / 0.13 & $\;\,\times\;$ / 0.09 \\ \midrule
			\multirow{5}{*}{\rotatebox[origin=c]{90}{\parbox{1cm}{\centering Botanic Garden}}} & 1005-00 & \textbf{0.23} & 1.39 / 0.71 & 1.43 / 0.48 & $\;\,\times\;$ / 0.47 & 2.41 / \underline{0.28} \\
			& 1005-01 & \underline{0.17} & 0.47 / 0.56 & 0.55 / 0.37 & 0.38 / 0.37 & 1.13 / \textbf{0.13} \\
			& 1005-07 & \textbf{0.36} & 0.86 / 0.72 & 1.87 / 0.69 & 0.55 / \underline{0.47} & $\;\,\times\;$ / 0.71 \\
			& 1006-01 & \textbf{0.15} & 1.37 / 1.15 & 0.46 / 1.23 & $\;\,\times\;$ / 0.44 & 4.76 / \underline{0.29} \\
			& 1008-03 & \textbf{0.24} & 0.52 / 0.74 & 0.82 / 0.43 & $\;\,\times\;$ / \underline{0.30} & 2.03 / 0.54 \\ \midrule
			\multirow{8}{*}{\rotatebox[origin=c]{90}{GRACO}} & aerial-01 & \textbf{0.20} & $\;\,\times\;$ / $\;\,\times\;$ & $\;\,\times\;$ / $\;\,\times\;$ & $\;\,\times\;$ / $\;\,\times\;$ & $\;\,\times\;$ / \underline{0.32} \\
			& aerial-02 & \textbf{0.07} & $\;\,\times\;$ / 0.25 & 1.67 / $\;\,\times\;$ & $\;\,\times\;$ / $\;\,\times\;$ & $\;\,\times\;$ / \underline{0.09} \\
			& aerial-03 & \textbf{0.07} & $\;\,\times\;$ / $\;\,\times\;$ & $\;\,\times\;$ / $\;\,\times\;$ & $\;\,\times\;$ / $\;\,\times\;$ & $\;\,\times\;$ / \underline{0.18} \\
			& aerial-04 & \textbf{0.12} & $\;\,\times\;$ / $\;\,\times\;$ & $\;\,\times\;$ / $\;\,\times\;$ & $\;\,\times\;$ / $\;\,\times\;$ & $\;\,\times\;$ / \underline{0.35} \\
			& aerial-05 & \textbf{0.21} & $\;\,\times\;$ / $\;\,\times\;$ & $\;\,\times\;$ / $\;\,\times\;$ & $\;\,\times\;$ / $\;\,\times\;$ & $\;\,\times\;$ / \underline{1.77} \\
			& aerial-06 & \textbf{0.07} & $\;\,\times\;$ / \underline{0.14} & 1.68 / 1.49 & $\;\,\times\;$ / $\;\,\times\;$ & $\;\,\times\;$ / 1.09 \\
			& aerial-07 & \textbf{0.09} & $\;\,\times\;$ / 6.71 & 1.59 / 1.74 & $\;\,\times\;$ / $\;\,\times\;$ & $\;\,\times\;$ / \underline{1.00} \\
			& aerial-08 & \textbf{0.13} & $\;\,\times\;$ / 0.24 & 1.00 / 1.93 & $\;\,\times\;$ / $\;\,\times\;$ & $\;\,\times\;$ / \underline{0.21} \\ \midrule
			\multirow{8}{*}{\rotatebox[origin=c]{90}{GEODE}} & water-short-alpha & \textbf{0.54} & $\;\,\times\;$ / \underline{2.56} & $\;\,\times\;$ / $\;\,\times\;$ & $\;\,\times\;$ / $\;\,\times\;$ & $\;\,\times\;$ / $\;\,\times\;$ \\
			& water-short-beta & \textbf{0.26} & $\;\,\times\;$ / 0.64 & $\;\,\times\;$ / 0.35 & 4.52 / 0.85 & $\;\,\times\;$ / \underline{0.33} \\
			& water-short-gamma & \textbf{0.27} & 0.38 / 0.36 & 0.30 / \underline{0.28} & $\;\,\times\;$ / $\;\,\times\;$ & $\;\,\times\;$ / \textbf{0.27} \\
			& offroad1-alpha & \textbf{0.11} & $\;\,\times\;$ / $\;\,\times\;$ & 1.66 / \underline{0.12} & $\;\,\times\;$ / 0.18 & $\;\,\times\;$ / \textbf{0.11} \\
			& offroad1-beta & \textbf{0.14} & 0.30 / 0.23 & 5.88 / 0.20 & $\;\,\times\;$ / \textbf{0.14} & $\;\,\times\;$ / \underline{0.15} \\
			& offroad1-gamma & \underline{0.13} & 1.71 / 0.30 & 6.14 / 0.24 & $\;\,\times\;$ / $\;\,\times\;$ & $\;\,\times\;$ / \textbf{0.12} \\
			& stairs-alpha & \textbf{0.23} & $\;\,\times\;$ / \underline{8.63} & $\;\,\times\;$ / $\;\,\times\;$ & $\;\,\times\;$ / $\;\,\times\;$ & $\;\,\times\;$ / $\;\,\times\;$ \\
			& stairs-beta & \textbf{0.09} & 0.92 / 0.24 & $\;\,\times\;$ / 0.28 & 5.62 / $\;\,\times\;$ & $\;\,\times\;$ / \underline{0.21} \\
			& tunnel3-alpha & \underline{0.20} & \textbf{0.18} / \textbf{0.18} & 1.22 / \textbf{0.18} & \textbf{0.18} / 1.26 & $\;\,\times\;$ / \textbf{0.18} \\
			& tunnel3-beta & \textbf{0.15} & 0.19 / 0.19 & 1.80 / 0.20 & $\;\,\times\;$ / 0.20 & \underline{0.16} / \textbf{0.15} \\
			& tunnel4-gamma & \textbf{0.17} & 2.33 / 4.44 & 0.90 / 0.31 & 0.61 / $\;\,\times\;$ & \underline{0.19} / \underline{0.19} \\ \bottomrule 
		\end{tabular}%
	}
	\begin{tablenotes}[flushleft]
		\item Table 1.\quad Results for EllipseLIO (ours) and the compared \gls{lio} approaches on five datasets. The best results are marked in \textbf{bold} and the second-best are \underline{underlined}. Experiments with APE RMSE errors of $\ge 10\,$m diverged and are marked with an $\times$.
	\end{tablenotes}
	\vspace{-2mm}
\end{table}  
	
	\begin{table}[tpb]
		\tbllabel{compute}
		\centering
		\scriptsize
		\adjustbox{width=\columnwidth,center}{%
			\begin{tabular}{@{}lccccc@{}}
				\toprule
				& EllipseLIO & DLIO & FAST-LIO2 & LIO-SAM & iG-LIO \\
				Voxel Resolution (m) & 0.1 & 0.1 / 0.25 & 0.1 / 0.5 & 0.1 / 0.4 & 0.1 / 0.5 \\ \midrule
				Computation Time (ms) & 35 & \underline{33} / 34 & \textcolor{gray}{203} / 51 & \textcolor{gray}{157} / 91 & \textcolor{gray}{31} / \textbf{11} \\
				Memory Usage (GB) & 2.4 & 7.2 / 2.9 & \textcolor{gray}{4.3} / \underline{0.7} & \textcolor{gray}{2.2} / \underline{0.7} & \textcolor{gray}{6.9} / \textbf{0.5} \\ \bottomrule
			\end{tabular}%
		}
		\begin{tablenotes}[flushleft]
			\item Table 2.\quad Mean computation time and memory usage for all of the \gls{lio} approaches on the \textit{parkland-mound} sequence. Approaches that diverged are in \textcolor{gray}{gray} and excluded.
		\end{tablenotes}
	\end{table}  

	The odometry results (Table 1) show that EllipseLIO is the best-performing approach overall. It consistently attains the lowest or second-lowest odometry error across all of the dataset sequences and achieves a $35\%$ lower \gls{ape} on average than the second-best performing approach, iG-LIO $0.5\,$m. EllipseLIO achieves a $68\%$ lower \gls{ape} on  average than the second-best approach with a $0.1\,$m voxel resolution, DLIO $0.1\,$m. EllipseLIO is also the only approach that runs on every dataset sequence without diverging.        
	
	EllipseLIO is able to provide accurate odometry estimates for all of the sequences regardless of their structure by adapting the scan registration metrics used to the local surface geometry. The ellipsoid representation captures the saliency of different geometric primitives \figref{primitives}: in structured environments (e.g., \textit{observatory-01} and \textit{aerial-04}) over $70\%$ of the ellipsoids are primarily planar, while in natural environments (e.g., \textit{parkland-mound} and \textit{1006-01}) these account for less than $55\%$. The saliency of these primitives is highly stable even when using lower resolution sensors: for  \textit{observatory-01} and \textit{parkland-mound}, which use 64-line LiDARs, only $5\%$ of the ellipsoids change their primitive classification, while for \textit{1006-01} and \textit{aerial-04}, which use 16-line LiDARs, this increases to $10\%$ and $13\%$ respectively.
	   
	EllipseLIO is able to correct drift when the sensor returns to a previously visited location by adaptively weighting point matches during the scan registration. This is shown by the close alignment between the EllipseLIO and ground truth trajectories for the \emph{bodleian-02} sequence \figref{bodleian}, while the compared approaches exhibit greater drift.
	
	The runtime performance of EllipseLIO and the compared approaches is evaluated on the \textit{parkland-mound} sequence as this produces a dense pointcloud map (Table 2). iG-LIO $0.5\,$m has the lowest computation time and memory usage. EllipseLIO runs $3\times$ slower and uses $5\times$ more memory to create a map with a $5\times$ higher voxel resolution. DLIO is the only compared approach that runs on this sequence with a $0.1\,$m voxel resolution without diverging. It has a similar computation time to EllipseLIO and uses $3\times$ more memory.
	
	\subsection{Ablation Study}

	Results from an ablation study demonstrate the impact of key components on the performance of EllipseLIO (Table 3). The range-based filtering approach is replaced with uniform filtering at the map resolution, which provides the same configuration as the compared \gls{lio} approaches by matching the scan filtering and map resolutions. This increases the mean APE over the successful sequences by 3$\times$. It causes divergence on many of the \emph{aerial-*} and \emph{water-*} sequences as uniform filtering does not account for the increased separation between scan lines at longer distances.
	
	 	\begin{figure}[tpb]
		\centering
		\subfloat[]{\includegraphics[width=0.1\textwidth]{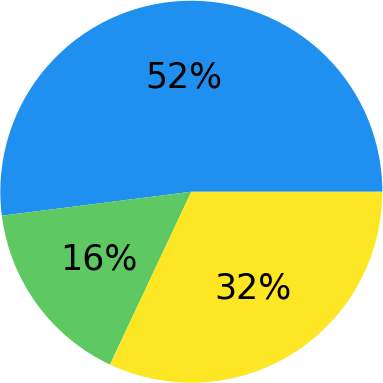}}\hfill
		\subfloat[]{\includegraphics[width=0.1\textwidth]{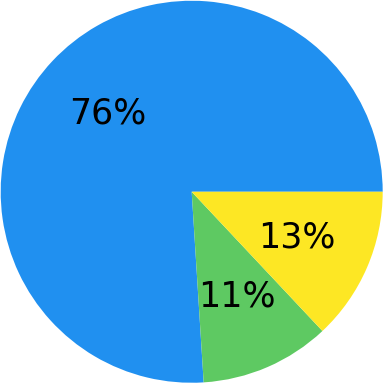}}\hfill
		\subfloat[]{\includegraphics[width=0.1\textwidth]{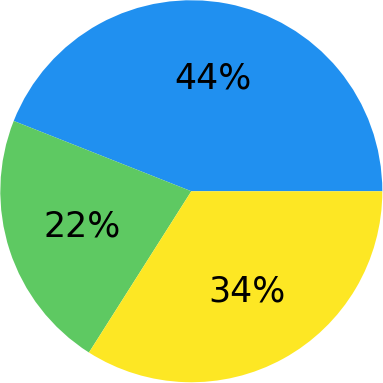}}\hfill
		\subfloat[]{\includegraphics[width=0.1\textwidth]{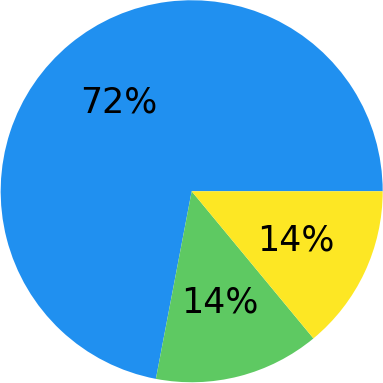}}
	\caption{The distribution of ellipsoid classifications, as planes (blue), lines (green), and balls (yellow), for (a) \textit{parkland-mound}, (b) \textit{observatory-01}, (c) \textit{1006-01}, and (d) \textit{aerial-04}.}
		\figlabel{primitives}
		\vspace{-2mm}
	\end{figure}
	
	\begin{figure}[tpb]
		\centering
		\includegraphics[width=0.9\linewidth]{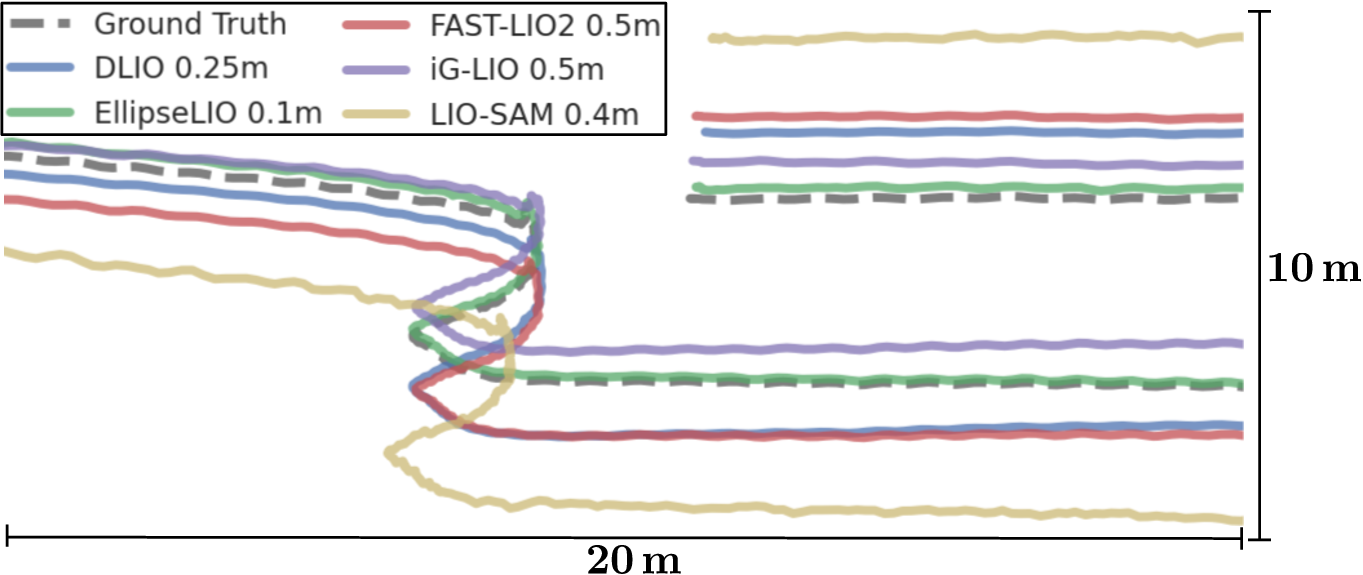}
		\caption{The \gls{lio} approach trajectories on the \textit{bodleian-02} sequence, showing their drift when aligned with the ground truth. EllipseLIO has the lowest drift due to the corrections provided by the adaptive match weighting.}
		\figlabel{bodleian}
		\vspace{-2mm}
	\end{figure}
	
	The saliency-weighted registration metrics are replaced with plane-to-plane GICP, point-to-plane, and point-to-point metrics. The plane-to-plane GICP metric obtains a surface estimate for scan points by computing and Eigen decomposing a covariance matrix from neighbouring scan points within a radius $r_i$ (Alg. 1, Line 4). Scan points with fewer than $n_\mathrm{min}$ neighbours have no surface estimate and use a point-to-plane metric. The plane-to-plane GICP and point-to-plane metrics produce the same mean APE as the full EllipseLIO for most of the sequences, but diverge in some moderately degenerate environments that provide poor planar constraints when using a lower-resolution VLP-16 or non-repeating Livox Avia (e.g., \emph{water-short-alpha}, \emph{water-short-gamma}, \emph{stairs-alpha}, and \emph{tunnel3-alpha}). The plane-to-plane GICP metric is more robust than the point-to-plane metric, but can only be used when there is a sufficient scan point density to obtain surface estimates. At longer ranges with lower-resolution LiDARs, the less robust point-to-plane metric must be used instead. 
	
	The point-to-point metric increases the mean APE for successful sequences by $56\%$ and diverges on \emph{tunnel3-alpha}, which demonstrates the importance of using planar constraints in structured environments. Replacing the adaptive match weighting with constant weights increases the mean APE for successful sequences by $63\%$ and causes divergence on \emph{water-short-gamma}, which illustrates the value of integrating drift correction into the scan registration.     

	\subsection{Sensitivity Analysis}

	Results from a sensitivity analysis (Table 4) show how changes to key parameters affect the performance of EllipseLIO and demonstrate that using the fixed values these parameters were originally assigned, across all of the experiments in Table 1, consistently provides the best overall performance. These key parameters are the map resolution, $\phi = 0.1$~m, the minimum number of points required to estimate an ellipsoid, $n_\mathrm{min} = 6$, and the maximum number of points that can be part of an ellipsoid, $n_\mathrm{max} = 60$. Setting $\phi = 0.25$~m only increases the mean APE for the successful sequences by $6\%$ but causes divergence on the \emph{stairs} sequence as the reduced point density produces ellipsoids with lower geometric fidelity. Setting $\phi = 0.5$~m increases the mean APE for the successful sequences by $81\%$ and also causes divergence on the other \emph{stairs-*} sequences. 
	
	Setting $n_\mathrm{min} = 12$ only increases the mean APE by $13\%$ and causes no divergences as most map regions provide sufficient point density to satisfy this higher minimum value. However, using $n_\mathrm{min} = 24$ produces a $94\%$ increase in the mean APE for the successful sequences and causes divergence on many of the \emph{aerial-*} and \emph{water-*} sequences as these depend on long range measurements that do not provide sufficient point density. Setting $n_\mathrm{max} = 6$ prevents ellipsoids from being updated with new measurements. This increases the mean APE by $19\%$ and causes divergence on the \emph{stairs} sequence as creating small ellipsoids from few measurements without refinement produces lower fidelity surface geometry estimates. Setting $n_\mathrm{max} = 600$ has no significant effect on the odometry performance and only increases the memory usage (e.g., from $2.4$~GB to $3.4$~GB on \emph{parkland-mound}). 
	
	\begin{table}[tp]
		\centering
		\scriptsize
		\adjustbox{width=\columnwidth,center}{%
			\begin{tabular}{@{}cccccc@{}}
				\multicolumn{6}{c}{Ablation Results (Mean APE RMSE in meters / Divergence Count)} \\
				\toprule
				\shortstack{EllipseLIO\\(Full)} &
				\shortstack{Uniform\\Filtering} &
				\shortstack{Plane-to-Plane\\(GICP) Metric} &
				\shortstack{Point-to-Plane\\Metric} & 
				\shortstack{Point-to-Point\\Metric} &
				\shortstack{Constant\\Match Weights} \\ \midrule
				\textbf{0.16} / \textbf{0} & 0.52 / 9 & \textbf{0.16} / 2 & \textbf{0.16} / 3 & \underline{0.25} / \underline{1} & 0.26 / \underline{1} \\ \bottomrule
			\end{tabular}%
		}
		\tbllabel{ablation}
		\begin{tablenotes}[flushleft]
		\item Table 3.\quad Results for the full EllipseLIO and ablated variants on all of the sequences from Table 1. The APE values are computed from the successful sequences. 
		\end{tablenotes}
		\vspace{-2mm}
	\end{table}
	
	\begin{table}[tpb]
		\centering
		\scriptsize
		\adjustbox{width=\columnwidth,center}{%
			\begin{tabular}{@{}ccccccc@{}}
				\multicolumn{7}{c}{Sensitivity Results (Mean APE RMSE in meters / Divergence Count)} \\
				\toprule
				\multirow{2}{*}[-0.5em]{\shortstack{EllipseLIO\\(Full)}} & \multicolumn{2}{c}{$\phi$ (m)} & \multicolumn{2}{c}{$n_\mathrm{min}$} & \multicolumn{2}{c}{$n_\mathrm{max}$} \\ \cmidrule(lr{1em}){2-3} \cmidrule(lr{1em}){4-5} \cmidrule(l{1em}){6-7}
				& 0.25 & 0.50 & 12 & 24 & 6 & 600 \\ \midrule
				\textbf{0.16} / \textbf{0} & \underline{0.17} / \underline{1} & 0.29 / 3 & 0.18 / \textbf{0} & 0.31 / 5 & 0.19 / \underline{1} & \textbf{0.16} / \textbf{0} \\ \bottomrule
			\end{tabular}%
		}
		\label{sensitivity}
		\begin{tablenotes}[flushleft]
			\item Table 4.\quad Results for the full EllipseLIO and variants with parameter changes on the sequences from Table 1. The APE values are computed from the successful sequences. 
		\end{tablenotes}
		\vspace{-2mm}
	\end{table}
	
	\subsection{Limitations}
	
	EllipseLIO performs well in moderately degenerate environments, as shown in Table 1, but does not address the challenge of providing robust odometry in highly degenerate scenarios (e.g., long geometrically uniform corridors) and it may diverge in these circumstances. Providing reliable odometry in such scenarios may be possible by fusing visual measurements into the presented approach. EllipseLIO can be used in moderately large environments with reasonable memory usage, as shown by the results in Table 2, but when operating in city-scale environments it may be necessary to use a sub-mapping approach to satisfy memory constraints.       

	\section{Conclusion}
	
	EllipseLIO provides reliable odometry that can generalise across environments and LiDARs without scenario-specific tuning. Range-based filtering preserves more geometric information from LiDAR scans, ellipsoid-based scan registration adapts to local surface geometry and adaptive weighting of point matches is used to correct accumulated drift. Experiments on five challenging datasets show that EllipseLIO achieves at least $24\%$ lower average \gls{ape} than compared \gls{lio} approaches on every dataset and does not diverge. Future work will integrate visual information into EllipseLIO to provide reliable odometry in degenerate environments.
	
	\renewcommand*{\bibfont}{\scriptsize}
	{\renewcommand{\markboth}[2]{}
		\renewcommand*{\UrlFont}{\rmfamily}
		\printbibliography}
	
\end{document}